# Towards Structuring an Arabic-English Machine-Readable Dictionary Using Parsing Expression Grammars

**Diaa M. Fayed[1] Aly Aly Fahmy[2] Mohsen Abdelrazek Rashwan[3] Wafaa Kamel Fayed[4]**

[1] Department of Computer Science, Faculty of Computers and Information,
Cairo University, diaa.fayed@outlook.com

[2] Department of Computer Science, Faculty of Computers and Information,
Cairo University, a.fahmy@fci-cu.edu.eg

[3] Department of EECE, Faculty of Engineering,
Cairo University, mrashwan@ieee.org

[4] Department of Arabic Language and Literatures, Faculty of Arts,
Cairo University., wafkamel@link.net

**Abstract**

*Dictionaries are rich sources of lexical information about words that is required for many applications of natural language processing and human language technology. However, publishers prepare printed dictionaries for human usage not for machine processing. This paper presented a method to structure partly a machine-readable version of the Arabic-English Al-Mawrid dictionary. The method converted the entries of Al-Mawrid from a stream of words and punctuation marks into hierarchical structures. The hierarchical structure expresses the components of each dictionary entry in explicit format. A dictionary entry is composed of subentries and each subentry consists of defining phrases, domain labels, cross-references, and translation equivalences. We designed the proposed method as cascaded steps where parsing is the main step. We implemented the parser using the parsing expression grammars formalism. In conclusion, although Arabic dictionaries do not have microstructure standardization, this study demonstrated that it is possible to structure them automatically or semi-automatically with plausible accuracy after inducing their microstructure.*

## 1. Introduction

Machine-readable dictionaries (MRDs) are electronic versions of the ordinary printed dictionaries. Publishers use MRDs to produce printed dictionaries. Fontenelle [1] differentiated between *computerized dictionaries* and *machine-readable dictionaries*. The computerized dictionaries are organized logically, so they are easily processed or transformed into lexical databases (LDBs) where each piece of information (definitions, parts-of-speech, grammar codes, etc.) is partly formalized and identified. The machine-readable dictionaries are flat text files with codes for driving the typesetting process. The computerized dictionaries have many applications [2] such as machine translation, natural language comprehension, pronunciation guides for speech recognition, etc.

Exploiting MRDs information by natural language processing (NLP) applications and research had two directions. The first direction is to structure the total text of a dictionary and to transform it into a computerized dictionary. The second direction is to extract a selected set of syntactic and semantic information and to use them, with other language sources such as dictionaries, encyclopedia, and corpora, to build a computational lexicon, a lexical taxonomy, a semantic network, etc. Examples of the first direction are [2-7] and examples of the second direction are [1, 8-14].

Although MRDs research is no longer active research for some main Latin languages such as English and French since researchers have shifted to machine-readable corpora [15, 16] as main language sources, MRDs research is still needed for resource-poor languages such as Arabic. The Arabic language still needs MRDs research for two purposes: to build language resources for Arabic NLP applications and to improve the Arabic lexicography in both the theoretical and industrial aspects. This study is the first published one to structure the Arabic-English Al-Mawrid dictionary automatically. It is worth mentioning that the Arabic WordNet [17-20] is a milestone project in the field of building open-source Arabic language resources.

The Al-Mawrid targets human user and has unstructured text format. Therefore, we need to convert the Al-Mawrid to a structured format before exploiting its data in any NLP applications. Generally, structuring or computerizing a dictionary text means that converting implicit information in the dictionary definitions into explicit information by identifying each significant piece of information and labeling it. We assume that the Al-Mawrid has large and rich lexical knowledge that is useful in NLP applications. Furthermore, we assume that the Al-Mawrid has a microstructure that we can capture and formalize automatically or semi-automatically. This study settled for the partly structuring of the Al-Mawrid definitions at the punctuation level without handling the scattered information in the defining phrases. The proposed method is composed of successive steps where parsing is the major step. We implemented the parser using the parsing expression grammars (PEGs) formalism.

Through investigating samples of the Al-Mawrid definitions, we found high regularity in their punctuations. Consequently, we assumed that parsing may be the more suitable technique for capturing the punctuation regularity. We adopted the PEGs formalism in this study for some specific advantages such as (a) the PEGs do not need a separate step of tokenization since tokenization rules take the same formats of any other grammar rule; and (b) the PEGs have implemented in a Python library and the Python programming language has high performance and flexibility in text processing. Furthermore, the PEGs have some other advantages that make it distinguishable. For more deep study of the PEGs, you can refer to [21].

The contributions of this paper are fourfold:

1. The first published study to computationally analyze and computerize the Al-Mawrid Arabic-English dictionary.
2. The first time to use the PEGs to parse dictionary entries.
3. The study reveals some shortcomings in the Al-Mawrid from a computational aspect.
4. The study reveals linguistic information in the Al-Mawrid that is useful for NLP applications.

The rest of the paper is organized as follows. Section 2 introduces a short review of related works. Section 3 explains, in detail, the structure of Al-Mawrid structure and types of information in its definitions. Section 4 defines the formal definition of the PEGs formalism. Section 5 describes the steps of proposed structuring method and illustrative examples. Section 6 evaluates the overall structuring method. Finally, section 7 concludes the study and gives some future works.

## 2. Background and Related Works

Amsler [22] suggested that we need a database management system to enter the data of a dictionary, to edit text definitions and make semantic analysis, and to organize and store the resultant data of the analysis. The structuring task of a dictionary text is often compromised of three phases: (a) acquiring a machine-readable version of the dictionary and then loading its data into a database, (b) analyzing definitions, extracting information, and labeling them, and (c) adding the extracted knowledge to the database or building other forms of language resources such as lexical taxonomies, semantic networks, etc. Some studies merge the three phases such as Wilms [4] and others separate the phases such as Byrd [23].

MRDs researchers often used either one of two approaches to carry out the first phase of structuring a dictionary: (a) building a parser to structure the dictionary entries, and (b) using ad-hoc procedures to extract selected pieces of information, and then loading them into a database. In the following subsections, we give a short review of methods that are used to acquire and load the data of MRD into a database.

### 2.1 Building Parsers

Wilms [4] automated the Funk and Wagnalls Dictionary by building a parser based on pattern matching and transition network. The format of the parser output is Prolog clauses, where parentheses enclosed fields. Byrd [23] built a tree structure for each entry across three stages. The first stage segments information on the dictionary tape to separated entries and stores them in a direct access file. The keys in the file are the headwords and the data are the bodies of the entries. The second stage identifies and labels each kind of data in an entry to consistent and standard format. Byrd developed a parser to transform each entry of the dictionary into a tree structure. In Nagao *et al*. [3], a data entry entered the printed versions of the dictionaries as it is. Then, Nagao built an augmented transition network (ATN) parser to transform the implicit relationships that are represented by special symbols or character types into explicit relationships that are represented by formatted records and links or pointers.

### 2.2 Developing Ad-hoc procedures

Nakramura and Nagao [24] extracted specific fields of the magnetic tape of the LDOCE and loaded them into a relational database. They developed extraction programs with a pattern matching based algorithm to analyze the definition sentences in LDOCE. Soha [25] transformed the unstructured text files of the RDI dictionary into a relational database. Soha carried out the transformation in three phases: (a) analyzing the text definitions, (b) imposing adopted structure on the RDI dictionary, and (c) splitting up the text definitions using punctuation and filling a relational database. Fontenelle [1] transformed the typesetting taps of Collins-Robert French-English dictionary into a relational database. Fontenelle wrote a series of software applications for retrieving records and a program for enriching the database with two fields. Alshawi [5] used a series of batch jobs and a general text editor to transform the characters stream of the LDOCE data structures on typesetting taps into Lisp s-expressions (symbolic expressions). Mayfield and McNamee [6] developed a script language called ALBET. Kazman [2] transformed the Oxford English Dictionary (OED) and its supplement into a MRD using manual system of mark-up and data entry.

## 3. The Al-Mawrid Structure

The Arabic-English Al-Mawrid dictionary is a general-purpose dictionary. The Al-Mawrid is a medium-sized dictionary that has approximately 35000 distinct headwords and about 60000 subentries. Investigating and understanding the structure of a dictionary entails understanding its macrostructure, its microstructure, and its information content in definitions. In Hartman [26], the macrostructure refers to the overall access format of the dictionary. In addition, the microstructure refers to how an entry is structured, how the information about a headword is provided and presented, and how the discourse structure of an entry is prepared appropriately for benefit of the anticipated user.

The author of The Al-Mawrid arranged the headwords alphabetically according to the first letters of the headwords. An entry of The Al-Mawrid starts by a bold headword. When a headword has more than one meaning or sense, each meaning occupies a subentry that is cited in separated lines. Subentries that express collocations, idioms, or examples are cited lately. A subentry consists of a mandatory Arabic section and an optional English section (the English translation). In addition, for distinguishing the headwords, each subentry text, except the first one, is shifted from the headword by two space characters. The Arabic section consists of three fields that we name *header*, *explanation* and *cross-reference*. The first field of an Arabic section is mandatory but the other two fields are optional. A colon separates the header from its explanation. A dash may precede the cross-reference field. The subentry, and consequently, their defining phrases or translation equivalence (TE) phrases may express *declaration*, *question*, or *exclamation*.

The three fields of an Arabic section have the same structure: each field consists of one or more words or phrases that are separated by either the Arabic comma or the conjunction word “أو”. These comma-separated phrases are almost synonymous or near synonyms.

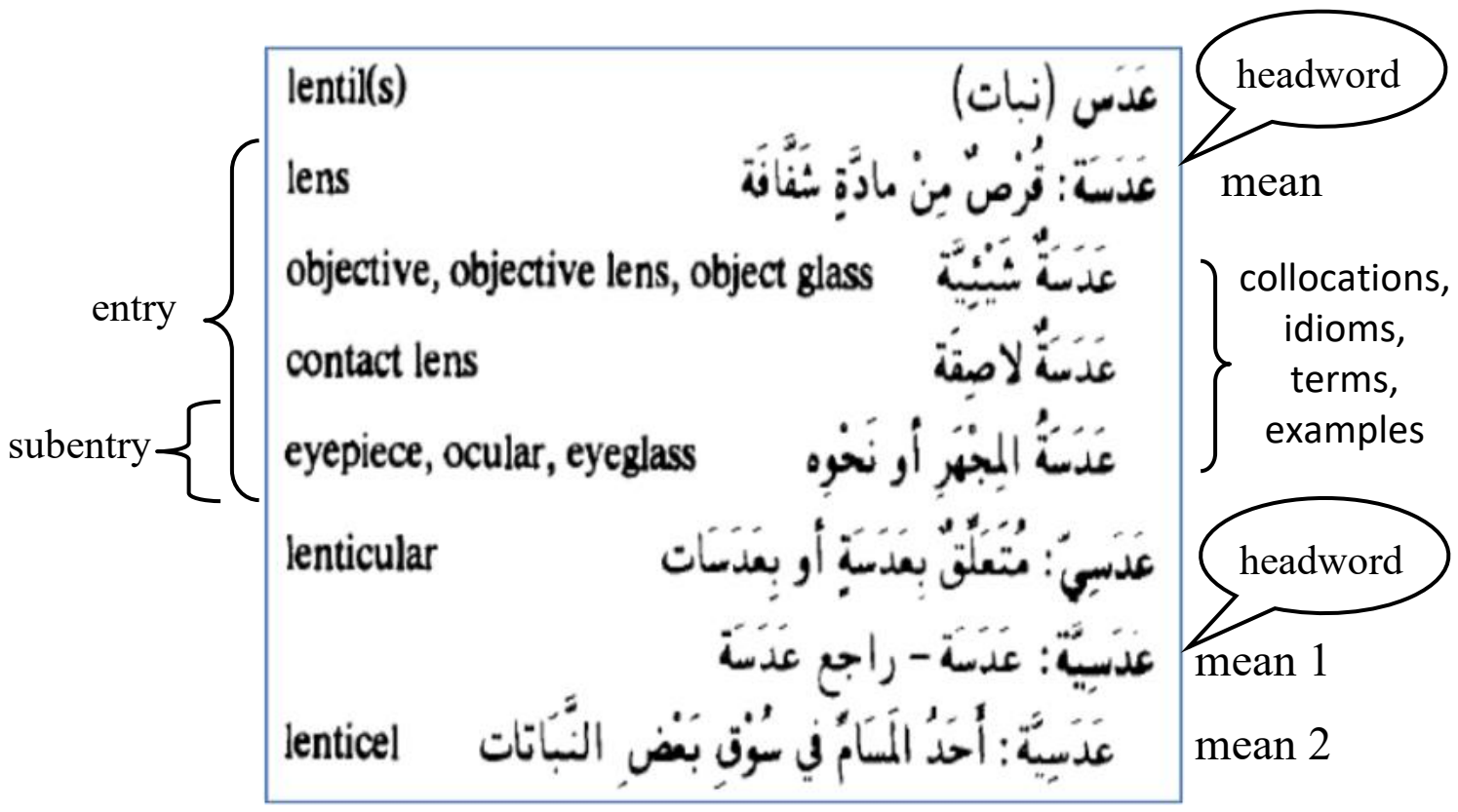

**Figure 1**: Some entries of the AL-Mawrid

The header has the headword of an entry if its subentry represents a sense of the headword. If a subentry does not represent a sense of the headword, it contains idioms, examples or restricted collocation. The morphologic, syntactic, or semantic information is scattered in the header or explanation fields. Some semantic domain or subject labels, such as “نبات” (plant), are enclosed in bracket parentheses (e.g. [نبات]) or in parentheses (e.g. (نبات)).

The English section has one or more translation equivalence groups (TEGs) that are separated by semicolons. Each TEG has one or more translation equivalence (TE) phrases that are separated by comma. The phrases of a TEG are synonyms. Figure 1 shows the entry structure of the headword “مُتَقَدِّم” in the printed version of the Al-Mawrid.

Because the author of the Al-Mawrid use synonyms extensively as a device of definitions , we will elaborate on them. Table 1 contains some meanings of the headword “عَيْن”. The synonyms in both Arabic and English sections may by single words or phrases that are composed of multiple words. Examples of multiple word synonym are prepositional phrases. Consider four meanings (or senses) of the noun headword “عَيْن”. the words "مُسْتَطْلِع" and "مُسْتَكْشِف" are synonyms and they also are synonyms to "عَيْن". Also, the words "scout", "reconnoiterer", and "explorer" are synonyms in mean 2, the words "مُرَاقِب" and "ناظِر" are synonyms and they also are synonyms to "عَيْن" in mean 3, the word "حارس" is synonyms to "عَيْن" in mean 4, the phrase "عُضْوُ البَصَر" is defining phrase. In conclusion,
in the first three sub-entries, the lexicographer used the synonyms as the main device for definition. on the other hand, in the fourth sub-entry, the lexicographer used defining phrases that composed of "genus" (عُضْوُ) and "differentia" (البَصَر).

| headword | Arabic | English |
|---|---|---|
| | … | … |
| عَيْن | عَيْن: مُسْتَكْشِف، مُسْتَطْلِع | scout, reconnoiterer, explorer |
| | عَيْن: ناظِر، مُرَاقِب | overseer, inspector, supervisor, superintendent |
| | عَيْن: حارس | lookout, guard, watch(man) |
| | عَيْن: عُضْوُ البَصَر | eye |
| | … | … |

**Table 1:** samples illustrates the synonyms in the Al-Mawrid

Arabic dictionaries have problems with some aspects that lexicographers should solve [27]. These problems are challenges for computerizing Arabic dictionaries. Hawaary [27] mentioned some deficiencies in the Arabic dictionaries at three levels: the theoretical foundation, the analysis, and the representation of syntactic-semantic content. In addition, The Al-Mawrid has some special deficiencies. The following issues are some of the shortcomings in the Al-Mawrid that complicate the structuring or the computerizing tasks.

1. *Inconsistency*:
   a. Semantic domain names or subject labels exist either in bracket parentheses [نبات] or in parentheses (نبات). A semantic relation may have more than one key word that manifests it. For example, the antonym relation is expressed by any of the words “ضد”, “عكس”, and “لا”.
   b. The adjective “صفة” and name “اسم” are the only part-of-speech tags in the Al-Mawrid and they are very rare.
   c. There is no a consistent clue that determines whether a subentry is either a meaning of the headword or an example, a collocation, an idiom, etc. A colon mark distinguishes the meaning only if it exists after a copy of the headword.
   d. The explanation in a subentry may be expressed by phrases that follow a colon mark, by phrases in parenthesis like “عقيد (في الجيش)”, or by nothing at all.

2. *Ambiguity*: A cross-reference word or phrase may refer to more than one headword and each headword may have more than one subentry. Furthermore, a subentry may represent either a meaning of the headword or an example, an idiom, a term, and a restricted collocation.
3. *Parentheses*: The parentheses information has large varieties and has multiple purposes of functions. For example, the parentheses may contain prepositions, synonyms, examples, explanations, antonyms, morphological information, spelling variations, etc.
4. *Indefinite key phrases*: The author of the Al-Mawrid does not cite key phrases or key words that manifest morphologic, syntactic, or semantic information in the introduction of the Al-Mawrid, so we need to discover these words before either structuring definitions or extracting information.

## 4. Parsing Expression Grammars

Parsing Expression Grammars (PEGs) formalism was built on the Backus-Naur Forms (BNFs) [28] and the Regular Expressions (REs) notations to form an alternative class of formal grammar. The PEGs formalism is essentially analytic rather than generative[1]. A PEG is defined [21] formally as a 4-tuple $G = (V_N, V_T, R, e_S)$, where $V_N$ is a finite set of nonterminal symbols, $V_T$ is a finite set of terminal symbols, $R$ is a finite set of rules, $e_S$ is a parsing expression termed the *start expression*, and $V_N \cap V_T = \mathcal{E}$. Each rule $r \in R$ is a pair $(A, e)$, which we write $A \leftarrow e$, where $A \in V_N$ and $e$ is a parsing expression. For any nonterminal $A$, there is exactly one $e$ such that $A \leftarrow e \in R$. $R$ is therefore a function from nonterminal symbols to expressions, and we write $R(A)$ to denote the unique expression $e$ such that $A \leftarrow e \in R$.

Consider $e$, $e_1$, and $e_2$ are parsing expressions. Furthermore, The variables $a$, $b$, $c$, $d$ are used to represent terminals, $A$, $B$, $C$, $D$ for nonterminals, $x$, $y$, $z$ for strings of terminals, and $e$ for parsing expressions. Inductively, *parsing expressions* are defined as follows [21]:

1. $\mathcal{E}$, the empty string.
2. $a$, any terminal, where $a \in V_T$ .
3. $A$, any nonterminal, where $A \in V_N$.
4. $e_1e_2$, a sequence.
5. $e_1/e_2$, prioritized choice.
6. $e^*$, zero-or-more repetitions.
7. $!e$, a not-predicate.

We used the Pyparsing library [2] that is a python implementation of the Parsing Expression Grammar [29, 30].

## 5. Structuring Method

Though the punctuation of the Al-Mawrid definitions has high percentage of regularity that we can capture and formalize, the morphologic, syntactic, and semantic information are scattered in the defining phrases randomly and inconsistently. Therefore, our strategy to formalize an entry of the Al-Mawrid is to capture the punctuation regularity by using a parsing technique firstly, and then to extract the scattered information by using an information extraction technique. This *divide-and-conquer* strategy avoids building complex grammars in order to capture both punctuations and scattered information in one shot.

Our current work addressed only the first phase of structuring dictionary text: acquiring and loading the data of the Al-Mawrid into a database format. We cannot structure the total text of a large dictionary like the Al-Mawrid in one automatic shot, but we often accomplish this by some combination of automatic, semiautomatic, and manual steps. We subdivided the structuring task into cascaded steps and the output of each step is the input to the consequent step. We implemented each step by one module. The modular design makes both testing code and evaluating step easier. Another advantage of the modular design is to avoid building complex and convoluted grammar for the Al-Mawrid that can take much time and effort. The complex grammar of the Al-Mawrid definitions results from the inconsistency and rich variability formats of definitions (especially parentheses information). In the following section, we explain in detail the structuring steps of the proposed method. Figure 2 summarizes the structuring method.

[1] *Wikipedia, Backus–Naur Form. 2013; Available from: http://en.wikipedia.org/wiki/Backus%E2%80%93Naur_Form.*
[2] McGuire, P. *Pyparsing Wiki Home*. 2013; Available from: http://pyparsing.wikispaces.com/home.

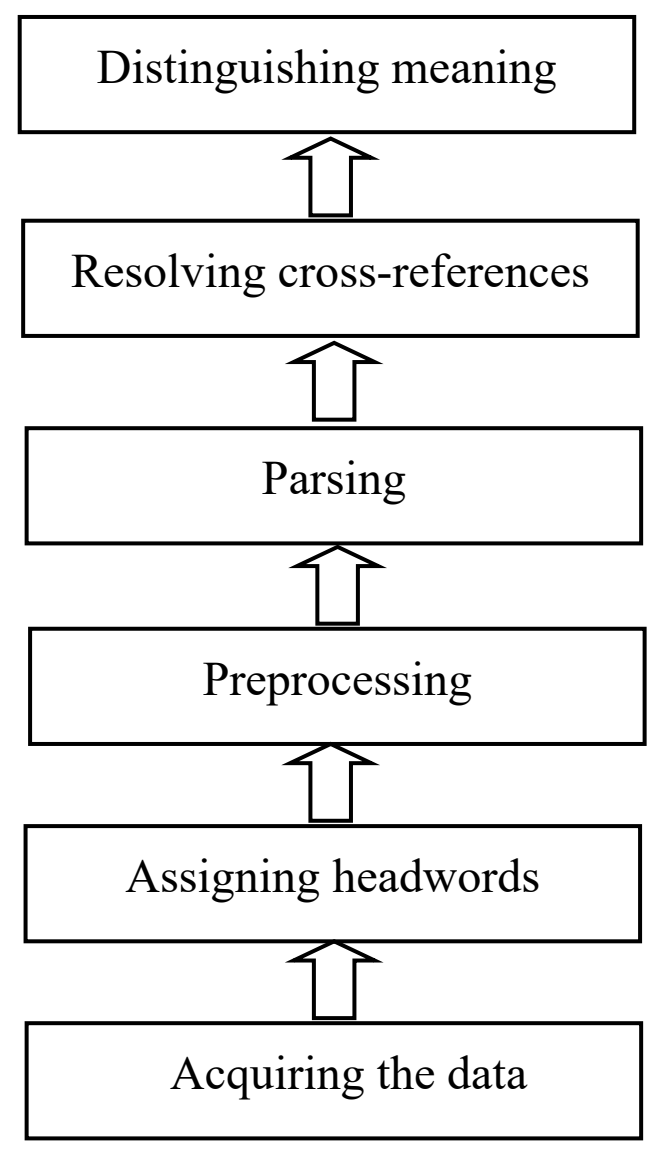

**Figure 2**: Steps of the structuring method

### 5.1 Acquiring the Dictionary Data

A human data entry explored the software in the Al-Mawrid CD and copied the entries information of headwords to a spreadsheet file. The data entry dropped out some information such as pictures and font styles. Each subentry of a headword occupies two separated rows in the spreadsheet file, one for Arabic section and the other for the English translation. Two reasons to transfer the data from the Al-Mawrid's CD to a spreadsheet file: (a) the spreadsheet is more suitable file format to handle for the data entry; (b) the data automatically acquires some structure of the spreadsheet file, rows and columns structure. Table 2 contains a sample of the dictionary data after copied into a spreadsheet file.

| **headword** | **article** |
|---|---|
| عَدَس | عَدَس (نبات) |
| | lentil(s) |
| عَدَسَة: | عَدَسَة: قُرْصٌ مِنْ مادَّةٍ شَفَّافَة |
| | lens |
| | عَدَسَةٌ شَيْئِيَّة |
| | objective, objective lens, object glass |
| | عَدَسَةٌ لاصِقَة |
| | contact lens |
| | عَدَسَةُ المِجْهَرِ أو نَحْوِهِ |
| | eyepiece, ocular, eyeglass |
| عَدَسِيّ: | عَدَسِيّ: مُتَعَلِّقٌ بِعَدَسَةٍ أو بِعَدَسَات |
| | lenticular |
| عَدَسِيَّة: | عَدَسِيَّة: عَدَسَة ـ راجع عَدَسَة |
| | عَدَسِيَّة: أَحَدُ المَسَامِّ في سُوْقِ بَعْضِ النَّبَاتات |
| | lenticel |

**Table 2** Data in spreadsheet file

### 5.2 Assigning Headwords to their Description Data

The purpose of this step is to recover the principal database structure of the Al-Mawrid. The principal structure of a dictionary is composed of an index of headwords and the entries information that is related to these headwords. Recovering the principal database structure of a dictionary enables us to know the information associated with each headword separately. Consequently, we can make more exploring and processing on the associated information, collecting basic information about the dictionary entries, and assigning the results to the respective headwords. The associated information with a headword can be definitions, idioms, terms, collocation, or examples, and translations. We developed procedures to transform the spreadsheet file into a rudimentary database of two flat text files. In this rudimentary database, one file is the index file that contains the distinct headwords and index numbers that point to the associated information in a data file. The second file is a data file that contains the associated information of the headwords. Table 3 illustrates the two files of the rudimentary database structure.

| index | |
|---|---|
| headword | sense# |
| عَدَس | 35467 |
| عَدَسَة: | 35468 |
| | 35469 |
| | 35470 |
| | 35471 |
| عَدَسِيّ: | 35472 |
| عَدَسِيَّة: | 35473 |
| | 35474 |

| data | | | |
|---|---|---|---|
| sense# | word-form | Arabic | English |
| 35467 | عَدَس | عَدَس (نبات) | lentil(s) |
| 35468 | عَدَسَة: | عَدَسَة: قُرْصٌ مِنْ مادَّةٍ شَفَّافَة | lens |
| 35469 | | عَدَسَةٌ شَيْئِيَّة | objective, objective lens, object glass |
| 35470 | | عَدَسَةٌ لاصِقَة | contact lens |
| 35471 | | عَدَسَةُ المِجْهَرِ أو نَحْوِهِ | eyepiece, ocular, eyeglass |
| 35472 | عَدَسِيّ: | عَدَسِيّ: مُتَعَلِّقٌ بِعَدَسَةٍ أو بِعَدَسَات | lenticular |
| 35473 | عَدَسِيَّة: | عَدَسِيَّة: عَدَسَة ـ راجع عَدَسَة | |
| 35474 | | عَدَسِيَّة: أَحَدُ المَسَامِّ في سُوْقِ بَعْضِ النَّبَاتات | lenticel |

**Table 3:** assigning some of headwords to its articles

### 5.3 Preprocessing

In this step, we carried out three operations on the data: (a) cleaning the data from some useless words and faulted characters, (b) flattening or consuming some parentheses information, and (c) converting domain or subject labels in the parentheses to labels in square brackets. We removed some useless words and characters such as "...", "etc.", and "الخ". The Al-Mawrid uses parentheses abundantly and for diverse purposes. The volume and diversity of the parentheses information in the Al-Mawrid is one of many challenges to structure its data.

In the English section of a subentry, there are two kinds of parentheses: the *connected parentheses* and the *separated parentheses*. The connected parentheses express multiple word forms or morphological information in compressed form. An example of this kind is the string "susten(ta)tion." This a compressed form that expresses two alternative word forms: "sustentation" and "sustention". The second kind is the separated parentheses. An example of that kind is the string "to seek protection (in, with, from)". The separated parentheses have many variations and contain many kinds of information, so they need further study to cope with and to structure. We did not analyze separated parentheses in this study except the kind of *one-word* parenthesis.

The one-word parenthesis contains a set of continuous characters that expresses preposition, alternative word form, prefixes, etc. The one-word parentheses may be connected or separated. We structured all the one-word parentheses by one technique. We made two copies of the strings that contain parentheses: one copy without the parentheses information and another copy with parentheses information. Then, we concatenated the two copies of the strings and inserted a comma in between them. For example, we transform the string "the (Heavenly) Father" to "the Heavenly Father" and "the Father". This technique is based on a note of the author in the Al-Mawrid introduction: all the information in the parentheses is optional and can be removed.

In the Arabic section, all parentheses are of separated kind. Arabic parentheses contain various and multiple information, so they need more deep analysis before we can structure them. This study handled only parentheses that contain labels of semantic or subject domain like "**عائِق (نبات)**" that is transformed into " **عائِق[نبات]**". The purpose of this transformation is to normalize all semantic or subject labels to be in square bracket, so they can be captured and formalized in a PEG grammar.

### 5.4 Parsing

Since the author of the Al-Mawrid does not give a detail explanation about the structure of definitions, we should induce the grammar of the Al-Mawrid by investigating ideal examples. To induce the grammar of the microstructure, we used ideal selected examples of the Al-Mawrid data as an experiment sample. We built the grammars of the parser using the *bootstrapping* technique. The main output of the parser program is a tree structure Table 4 and Table 5 contains the grammars of Arabic microstructure and English microstructure. In addition, the parser program produces a by-product corpus of both Arabic defining phrases and English translation equivalences (TEs) phrases. We can use the resulting corpus in further analysis to discover string patterns that manifest additional morpho-syntactic and semantic information since we do not know, in advance, the sublanguage of the dictionary definitions, as explained above, we adopted the *data-driven* (bootstrapping) technique to build the parser. We built the grammars by investigating ideal examples of the dictionary bit by bit using the parsing expression grammars (PEGs). Figure 3 illustrates the methodology that is used to build the grammars. We stopped the bootstrapping process when the parser can parse all the dictionary entries. Figure 4 shows an example of an input and an output of the parser. Figure 5 illustrates a complete output of the parser and other steps of the structuring method.

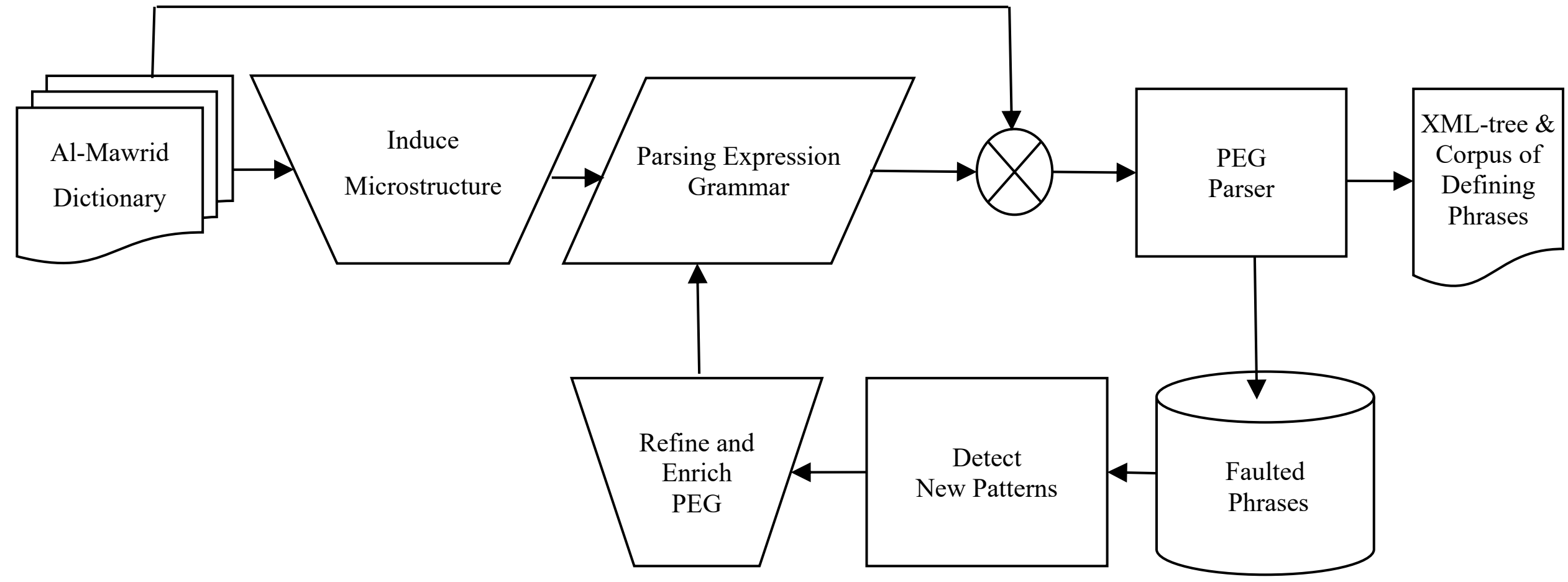

**Figure 3**: Bootstrapping the microstructure grammar

| **Hierarchical syntax** | |
|---|---|
| arabic | <- arabic1 / arabic2 / arabic3 / arabic4 / arabic5 / arabic6 |
| arabic1 | <- header domain? COLON domain? definition domain? (TATWEEEL / DASH) (REFERETHEM / REFERE) cross-reference |
| arabic2 | <- header domain? COLON domain? (TATWEEEL / DASH) (REFERETHEM / REFERE) cross_reference |
| arabic3 | <- header domain? COLON domain? definition domain? |
| arabic4 | <- header domain? (TATWEEEL / DASH) (REFERETHEM / REFERE) cross_reference |
| arabic5 | <- header domain? (REFERETHEM / REFERE) cross_reference |
| arabic6 | <- header domain? |
| header | <- sentence (ARABIC_COMMA  sentence)* |
| definition | <- sentence (ARABIC_COMMA  sentence)* |
| cross_reference | <- sentence (ARABIC_COMMA  sentence)* |
| sentence | <- (exclamation / exclamation / declaration)+ |
| declaration | <- simple_phrase parenthesis* simple_phrase* |
| question | <- (word)+ ARABIC_QUESTION_MARK |
| exclamation | <- (word)+ EXCLAMATION_MARK |
| simple_phrase | <- (!TATWELL !REFERE word)+ / parenthesis |
| word | <- [ARABIC_ALPHASLATIN_CHARAS]+ |
| parenthesis | <- LEFT_PARANTHESIS [ARABIC_ALPHAS LATIN_CHARAS]+ RIGHT_PARANTHESIS |
| domain | <- LEFT_SQUARE_BRACKET [^][]+ RIGHT_SQUARE_BRACKET |
| **# Lexical syntax** | |
| ARABIC_ALPHAS | <- 'ءاأآبتثجحدذرزسشصضطظعغفقكلمنهةويىًَُ ٌ ِ ٍ ْ ّ' |
| LATIN_CHARAS | <- '.\/0123456789' |
| REFERE | <- 'راجع' |
| REFERETHEM | <- 'راجعها' |
| LEFT_SQUARE_BRACKET | <- '[' |
| RIGHT_SQUARE_BRACKET | <- ']' |
| LEFT_PARANTHESIS | <- '(' |
| RIGHT_PARANTHESIS | <- ')' |
| DASH | <- '-' |
| TATWELL | <- 'ـ' |
| EXCLAMATION_MARK | <- '!' |
| ARABIC_QUESTION_MARK | <- '؟' |

**Table 4:** The grammar of Arabic microstruture

| **Hierarchical syntax** | |
|---|---|
| english | <- english1 \| english2 |
| english1 | <- header COLON definition |
| english2 | <- header |
| header | <- sense (SEMICOLON sense)* |
| definition | <- sense (SEMICOLON sense)* |
| sentence | <- (question \| exclamation \| declaration)+ |
| sense | <- sentence (ENGLISH_COMMA sentence)* |
| declaration | <- (word \| parenthesis)+ |
| question | <- (word)+ QUESTION_MARK |
| exclamation | <- (word)+ EXCLAMATION_MARK |
| word | <- [ENGLISH_ALPHASSPECIALS_CHARSDIGITS]+ |
| parenthesis | <- LEFT_PARANTHESIS [ENGLISH_ALPHAS SPECIALS_CHARS DIGITS]+ RIGHT_PARANTHESIS |
| **Lexical syntax** | |
| DIGITS | <- '123456789' |
| SPECIALS_CHARS | <- ',.-' |
| ENGLISH_ALPHAS | <- 'abcdefghijllmnopqrstuvwxyzABCDEFGHIJKLMNOPQRSTUVWXYZ'' |
| QUESTION_MARK' | <- '?' |
| EXCLAMATION_MARK | <- '!' |
| COMMA | <- ',' |
| SEMICOLON | <- ';' |
| COLON | <- ':' |
| LPARANTHESIS | <- '(' |
| RPARANTHESIS' | <- ')' |
| APOSTROPH | <- '\'' |

**Table 5** The grammar of English microstructure

Assembling the defining and translation phrases in a corpus enables us to process and analyze them by statistical methods to discover the string patterns that manifest morpho-syntactic and semantic relations. Table 6 shows examples of the defining phrases corpus.

| Definition samples | Type of Information | Corpus of defining phrases |
|---|---|---|
| (نبات)عَجْرَم | semantic domain | (نبات)عَجْرَم |
| “عَجَائِب(مفردها عَجِيبَة) ـ راجع عَجِيبَة” \| “عَجَائِب”<br>“عَجِيبَة(ج عَجَائِب): أُعْجُوبَة، مُعْجِزَة” \| “عَجِيبَة” | morphology | (مفردها عَجِيبَة)عَجَائِب<br>(ج عَجَائِب)عَجِيبَة |
| “عالِم(اسم)” \| “عالِم”<br>“عالِمٌ بِـ(صفة)” \| “عالِم” | syntactic:<br>part-of-speech | (اسم)عالِم<br>(صفة)عالِمٌ بِـ |
| " عارِض ": خَدّ، جانِبُ الوَجْه | definition | خَدّ<br>جانِبُ الوَجْه |
| " عاقّ ": عَقُوق، ضِدّ بارّ<br><br>خَيْلانِيّ: حَيَوانٌ مِنَ الخَيْلانِيَّات<br>لاذَنَبِيَّات: رُتْبَةٌ مِنَ البَرْمائِيَّات | Semantic relations: Antonyms<br><br>Hypernym/Hyponym relation | عَقُوق<br>ضِدّ بارّ<br><br>حَيَوانٌ مِنَ الخَيْلانِيَّات<br>رُتْبَةٌ مِنَ البَرْمائِيَّات |
| اِبْتِغاء: مَصْدَر اِبْتَغَى | Morphology: derivation | مَصْدَر اِبْتَغَى |
| " عارَكَ ": قاتَلَ، شاجَرَ<br>" قَرْض ": عارِيَة، عارِيَّة | Semantic relations: defining using synonyms, entry composed of synonyms | قاتَلَ<br>شاجَرَ<br>عارِيَة<br>عارِيَّة |
| قَفَّال: صانِعُ الأَقْفال | Semantic roles: Human/profession | |

**Table 6**: Examples of information in the defining phrases

### 5.5 Resolving Cross-References

The cross-reference in a dictionary is a mechanism that guides users to refer to other entries or subentries of the dictionary. The cross-reference in the Al-Mawrid refers to an entry or a subentry that is related to the given meaning or subentry.

The cross-reference device in Al-Mawrid is either the key word “راجع” (refer-to) or the key phrase “راجعها في أماكنها” (refer-to-them-in-their-place). The first refers explicitly to a word or a phrase after the key word “راجع” and the second specifies the referred words or phrases that the user should go before the key phrase “راجعها في أماكنها. The number of cross-references in the Al-Mawrid is about 13 % of the subentries. There are only about fifty of the second type in the Al-Mawrid.

#### 5.5.1 The Cross-reference Issues

The cross-references in the Al-Mawrid have some complexities: (a) a cross-reference may refer to one or more entries or subentries, (b) a cross-reference word may have more than one meaning, and (c) a cross-reference may be a phrase that is not a headword but that exists in a subentry as an idiom, a collocation, a verbal phrase, or an example. Therefore, we need to know which a headword we will search the dictionary to get the meanings of that phrase. Figure 4 explains the complexity of the cross-reference ambiguity. The headword “أَحَاطَ” has a subentry “أَحَاطَهُ بِالرِّعَايَةِ أو العِنَايَةِ أو الاهْتِمام ـ راجع رَعَى، عُنِيَ بِـ، اِهْتَمَّ بِـ”. The cross-reference phrases are “رَعَى”, “ عُنِيَ بِـ”, and “اِهْتَمَّ بِـ”. The cross-reference resolver search the index file to get the indices of the phrases: “رَعَى”, “عُنِيَ بِـ”, and “اِهْتَمَّ بِـ”. The word “رَعَى” is the only word that has indices numbers; the other two phrases return zero indices because they have prepositions. The index file contains only words (not phrases) as indices.

#### 5.5.2 Resolving the Cross-reference Ambiguities

We can resolve the cross-reference ambiguities in two steps. First, we replace the cross-reference words or phrases by a set of index numbers. Because of the ambiguity of some words, the set may have more than one number. Second, we identify exactly what sense is required, preserve its index number, and remove the remaining index numbers. Actually, we made the two steps only for the cross-reference key phrase “راجعها في أماكنها” because their size is small. On the other hand, the cross-reference key word “راجع” has large size, so we made only the first step for them to resolve ambiguity. The second step needs further study and analysis to solve ambiguities. Figure 6 is an example that illustrates the first step of resolving the cross-reference ambiguity. The cross-reference numbers is the “refs” value attribute of the crossreference element.

**Figure 4:** Example of cross-reference ambiguity

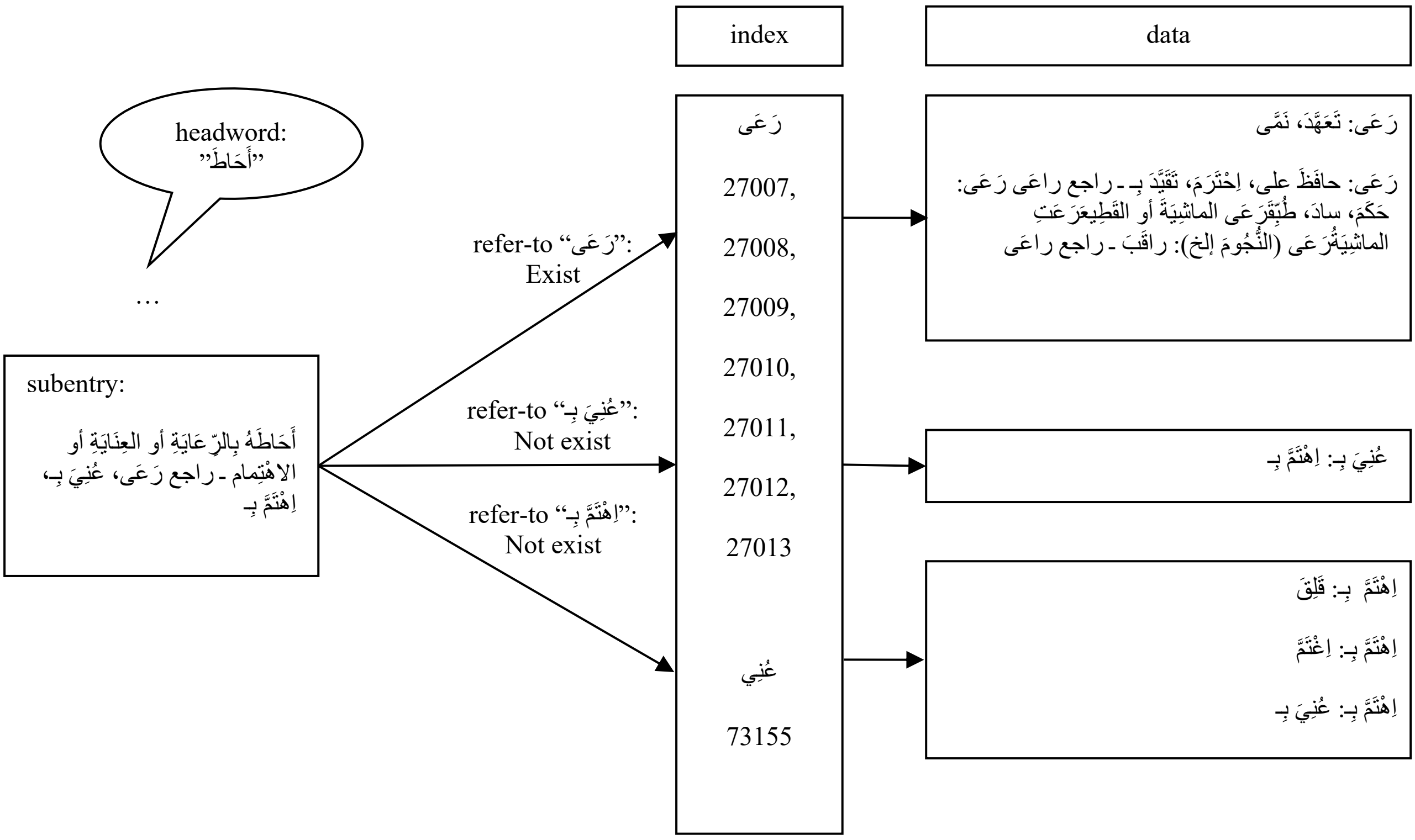

### 5.6 Distinguishing Meanings

The information that is associated with a headword may include meanings, idioms, terms, collocations, and examples. The Author of the Al-Mawrid cites the subentries of a headword without distinguishing between kinds of them. The author depends on the understanding of the dictionary user in distinguishing among all these kinds. However, structuring a dictionary entails making these kinds as distinguishable as possible.

We noticed that most of the subentries that contain meanings of a headword start by a copy of the headword itself and colon mark. We exploited this phenomenon by implementing a heuristic that acts as a distinguisher. We implemented the distinguisher as a flag that has a binary value. If a subentry represents a meaning of headword, the flag value equals to one. On the other hand, if a subentry represents an idiom, a term, a collocation, or an example, the flag value equals to zero. We implemented the flag as a binary attribute "ism" (is-meaning or is-sense) in the "sense" element of an entry sub-tree. Figure 7 illustrates two examples of the "ism" attribute. However, distinguishing among an idiom, a term, a collocation, or an example requires further knowledge and processing steps such as looking up an English dictionary. Figure 5 illustrates a complete output of the parser and other steps of the structuring method.

```xml
<entry id="19956" headword="عَدَسِيَّة">
 <group id="1">
  <sense id="35472" ism="1">
   <arabic>
     <header>
      <declaration>عَدَسِيَّة</declaration>
     </header>
     <definition>
      <declaration>عَدَسَة</declaration>
     </definition>
     <crossreference refs="35467,35468,35469,35470">
      <declaration>عَدَسَة</declaration>
     </crossreference>
   </arabic>
   <english/>
  </sense>
  <sense id="35473" ism="1">
   <arabic>
     <header>
      <declaration>عَدَسِيَّة</declaration>
     </header>
     <definition>
      <declaration>أَحَدُ المَسَامِّ في سُوقِ بَعْضِ النَّبَاتَات</declaration>
     </definition>
   </arabic>
   <english>
    <header>
     <subsense>
       <declaration>lenticel</declaration>
     </subsense>
    </header>
   </english>
  </sense>
 </group>
</entry>
```

```xml
<entry id="19953" headword="عَدَس">
 <group id="1">
   <sense id="35466" ism="1">
    <arabic>
      <header>
       <declaration>عَدَس</declaration>
      </header>
      <domain>[نبات]</domain>
    </arabic>
    <english>
      <header>
       <subsense>
         <declaration>lentils</declaration>
         <declaration>lentil</declaration>
       </subsense>
      </header>
    </english>
   </sense>
 </group>
</entry>
```

```xml
<entry id="19955" headword="عَدَسِيّ">
 <group id="1">
   <sense id="35471" ism="1">
    <arabic>
      <header>
       <declaration>عَدَسِيّ</declaration>
      </header>
      <definition>
       <declaration>مُتَعَلِّقٌ بِعَدَسَةٍ أو بِعَدَسَات</declaration>
      </definition>
    </arabic>
    <english>
     <header>
      <subsense>
       <declaration>lenticular</declaration>
      </subsense>
     </header>
    </english>
   </sense>
 </group>
</entry>
```

```xml
<entry id="19954" headword="عَدَسَة">
 <group id="1">
   <sense id="35467" ism="1">
    <arabic>
      <header>
       <declaration>عَدَسَة</declaration>
      </header>
      <definition>
       <declaration>قُرْصٌ مِنْ مادَّةٍ شَفَّافَة</declaration>
      </definition>
    </arabic>
    <english>
      <header>
       <subsense>
        <declaration>lens</declaration>
       </subsense>
      </header>
    </english>
   </sense>
   <sense id="35468" ism="0">
    <arabic>
      <header>
       <declaration>عَدَسَة شَيْئِيَّة</declaration>
      </header>
    </arabic>
    <english>
      <header>
       <subsense>
        <declaration>objective</declaration>
        <declaration>objective lens</declaration>
        <declaration>object glass</declaration>
       </subsense>
      </header>
    </english>
   </sense>
   <sense id="35469" ism="0">
    <arabic>
      <header>
       <declaration>عَدَسَة لاصِقَة</declaration>
      </header>
    </arabic>
    <english>
      <header>
       <subsense>
        <declaration>contact lens</declaration>
       </subsense>
      </header>
    </english>
   </sense>
    <sense id="35470" ism="0">
     <arabic>
       <header>
        <declaration>عَدَسَة المِجْهَر أو نَحْوِه</declaration>
       </header>
     </arabic>
     <english>
       <header>
        <subsense>
         <declaration>eyepiece</declaration>
         <declaration>ocular</declaration>
         <declaration>eyeglass</declaration>
        </subsense>
       </header>
     </english>
    </sense>
  </group>
</entry>
```

**Figure 5:** an output example of the strcutering steps

## 6. Results and Discussion

We selected the Ayn "ع" letter from the Al-Mawrid to evaluate the structuring steps. The Ayn "ع" letter represents about 4.6 % of the Al-Mawrid data volume. The only criteria for selecting the Ayn letter are that its volume is suitable for both the evaluation experiment and the copyright considerations. Table 7shows the sources of errors and errors count. Table 7 illustrates the types of errors for each source. From Figure 6, we inferred that the parser produced the smallest number of errors; this indicates that the entries of the Al-Mawrid have high-organized punctuation. We noticed also that the main sources of errors are the dictionary and the data entry. By analyzing the errors of each error source in Table 7, we noticed that the errors of cross-refs and is-meaning are duplicated errors of the first two error-sources: the in-dictionary and the data-entry. As a result, the main independent error-sources are the in-dictionary, the data-entry, and the parser. The cross-refs and is-meaning are dependent error-sources. If we compute only errors of the independent error-sources, the percentage of accuracy is approximately 91 %. On the other hand, if we take all errors of all error-sources, the accuracy is approximately 87%.

| Source of errors | Examples of errors |
|---|---|
| in-dictionary | ▪ The phrase in the cross-reference does not exist in the headword list.<br>▪ The colon that distinguishes meaning is absent.<br>▪ The head or cross-reference contains parentheses and/or comma.<br>▪ The head or cross-reference has diacritics.<br>▪ Entry meanings are shifted to another previous or followed entry.<br>▪ Mismatches exist between the electronic and the printed versions. |
| human data entry | ▪ Some entries and subentries are dropped out.<br>▪ Subentries are shifted to other entries.<br>▪ Mismatches exist between electronic and printed version. |
| parser | ▪ Faulted characters exist in the Arabic string.<br>▪ Spaces exist in a head phrase of subentry. |
| cross-reference (cross-ref) | ▪ Refs contain parenthesis.<br>▪ Refs are phrases or multi-words that do not exist in the index file.<br>▪ Some headwords are missed.<br>▪ Space is missed in a head phrase |
| is-sense (or is-meaning) | ▪ Header contains parenthesis.<br>▪ Refs contains parenthesis.<br>▪ Some headwords are absent.<br>▪ Some colons and comma are absent.<br>▪ Some headers are phrases.<br>▪ Header contains two or more words separated by comma.<br>▪ Refs contains preposition. |

**Table 7:** Source and thypes of errors

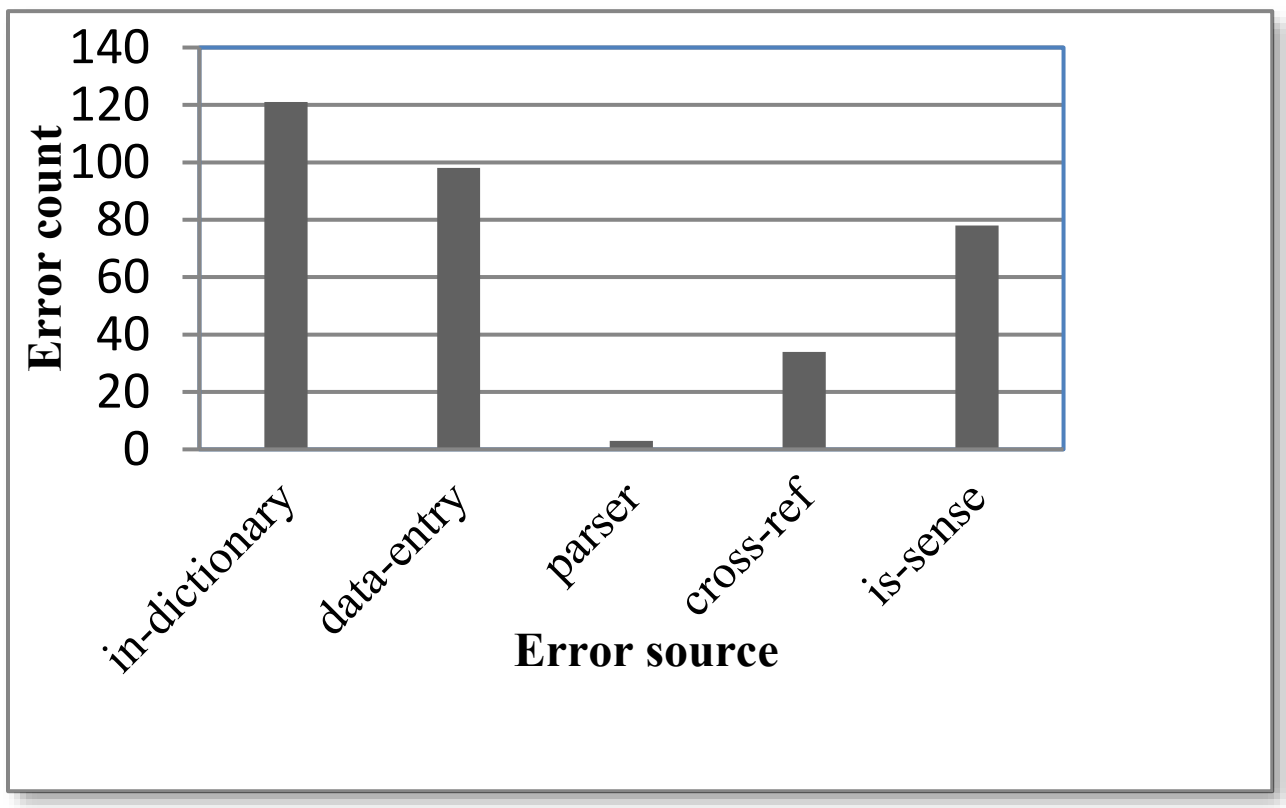

**Figure 6:** Error sources and their percentage

**Conclusions and Future Work**

The accuracy percentage of the structuring steps, especially of the parser step, proved our previous assumption; the Al-Mawrid has high punctuation regularity in its microstructure, and so it can be formalized automatically. In addition, investigating the by-product corpus of defining phrases showed that the defining phrases have prospective rich information that can be extracted and formalized. The main lesson of this study is that we can formalize Arabic dictionaries automatically, with plausible accuracy, thought they have no standardization formats. In future study, we can pursue the following tasks on the present work:

- Annotating the dictionary definitions by part-of-speech (POS) tags and Semantic tags,
- Analyzing and structuring the phrases in parentheses,
- Resolving the cross-references completely such that each cross-reference refers to the exact meaning of a specific headword,
- Extracting syntactic and semantic information from the corpus of defining phrases,
- Disambiguating words of the defining phrases and translation equivalence phrases,
- Analyzing the conjunction words in both Arabic "أو" and English "or".
- Transforming the resulted XML-tree of the Al-Mawrid into a standardized format like *Lexical markup framework (LMF) and TEI (Text Encoding Initiative) XML markup.*
- Designing and implementing a dictionary management tool to correct errors, enrich information, explore data, etc.

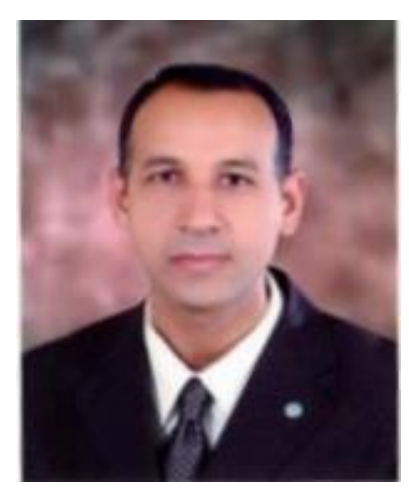

**Diaa El-Din Mohamed Abo-Fayed** received the B.Sc. degree in the Electronics, Faculty of Engineering, Mansoura University, 1995. He received M.Sc. degree in the Automatic Control from the Faculty of Engineering, Mansoura University, 2001, in the field of "Data Mining". Now Diaa is a PhD Candidate in the computer science in the Faculty of Computers and Information, Cairo University; the PhD is in the field of Arabic NLP. Diaa worked as an engineering research in the COLTEC ME company in building HLT for Arabic. Diaa also worked in the News Group Co. as an engineering research. The News Group Co. specialize in the sourcing, distribution, creation, monitoring and analysis of news content in the emerging markets of the Middle East, Africa and the Indian sub-continent. Currently, Diaa is an Electronics and Communications Engineer in the National Water Research Center.

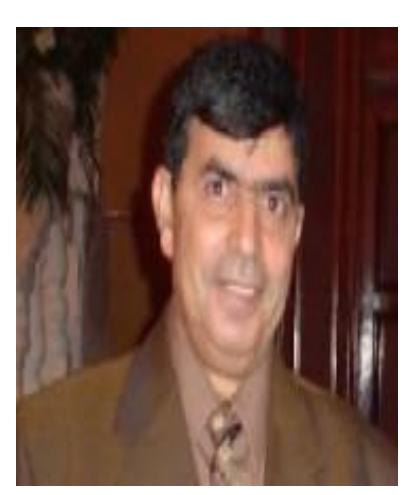

**Aly Aly Fahmy** is the former Dean of the Faculty of Computers and Information, Cairo University and a Professor of AI and ML, in the Department of Computer Science. He graduated from the Department of Computer Engineering, Technical College with honor degree. He specialized in Mathematical Logic. He received a master's degree from the ENSAE, Toulouse, France, 1976 in the field of Logical DB systems and then obtained his PhD from the Center for Studies and Research – CERT-DERI, 1979, Toulouse – France in the field of Artificial Intelligence. He participated in several national projects, built expert systems, and published many papers and books. Prof. Fahmy's main research areas are: Data and Text Mining, Mathematical Logic, Computational Linguistics, Text Understanding and Automatic Essay Scoring and Technologies of Man–Machine Interface in Arabic. He Directed the first Center of Excellence in Egypt in the field of Data Mining and Computer Modeling (DMCM). Prof. Aly Fahmy is currently involved in the implementation of the exploration project of Master's and Doctorate theses of Cairo University.

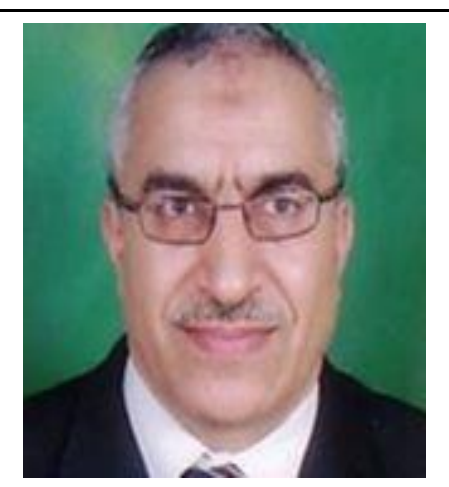

**Mohsen Abdelrazek Rashwan** received the B.Sc. and M.Sc. degrees in electronics and electrical communications from the Faculty of Engineering, Cairo University, Cairo, Egypt. The M.Sc. degree in systems and computer engineering from Carleton University, Ottawa, ON, Canada, and the Ph.D. degree in electronics and electrical communications from Queen's University, Kingston, ON, Canada. He currently serves as a Professor in the Department of Electronics and Electrical Communications, Faculty of Engineering, Cairo University, Cairo Egypt, and as the Managing Director of the Engineering Company for the development of Computer Systems that cofounded in 1993. Over the past research as well as building commercial products of Arabic NLP, digital speech processing, image processing, OCR, and e-learning. Among other several national and international mega projects, he has seved/been serving as a Senior Scientist in the EC's FP7 projects on Arabic HLT NEMLAR and MEDAR , and as a Co-PI in the Egyptian Data Mining and Computer Modeling Center of Excellence (DMCM-CoE).

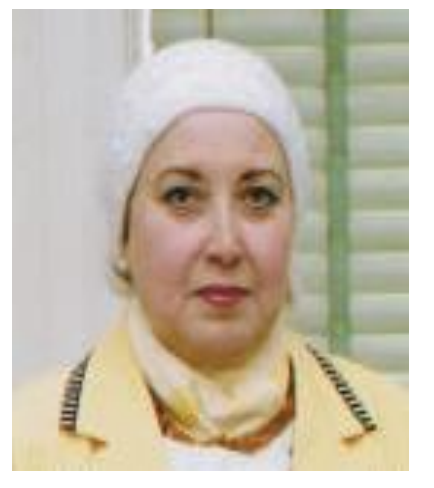

**Wafaa Kamel Fayed** B.A., M.A., Ph.D., Arabic Department, Faculty of Arts, Cairo University. Professor, Arabic Department - Faculty of Arts, Cairo University. Correspondent member, Arabic Language Academy, Damascus. Expert , Arabic Language Academy , Cairo. Supervisor of 49 Ph.D. and M.A. Degrees. Award of Ideal Professor at Cairo University. Honorary Certificate of distinction, 2004. Cairo University's Incentive award of human and social sciences, 2004. The 1st thesis using computer in applications of Arabic studies at 1974. Cairo University's appreciation award of humanities and pedagogic sciences 2013.
Published 58 academic researchers at universal journals. The author of 7 academic books and translator of 2 books.